\documentclass{ccr}

\title{A Computational Model of Message Sensation Value in Short Video Multimodal Features that Predicts Sensory and Behavioral Engagement}
\shorttitle{A Computational Model of MSV}
\authorsnames{Haoning Xue, Jingwen Zhang, Xiaohui Wang, Diane Dagyong Kim, Yunya Song}
\shortauthors{Xue, Zhang, Wang, Kim, \& Song}

\authorsaffiliations{
  {Department of Communication, University of Utah},
  {Department of Communication, University of California, Davis},
  {Department of Media and Communication, City University of Hong Kong},
  {Department of Communication, University of California, Davis},
  {Division of Emerging Interdisciplinary Areas, Hong Kong University of Science and Technology}
}

\abstract{
The contemporary media landscape is characterized by sensational short videos. While prior research examines individual multimodal features’ effects, the collective impact of multimodal features on viewer engagement with short videos remains unknown. Grounded in the theoretical framework of Message Sensation Value (MSV), this study develops and tests a computational model of MSV with multimodal feature analysis and human evaluation of 1,200 short videos. This model that predicts sensory and behavioral engagement was further validated across two unseen datasets from three short video platforms (combined \textit{N} = 14,492). While MSV is positively associated with sensory engagement, it shows an inverted U-shaped relationship with behavioral engagement: Higher MSV elicits stronger sensory stimulation, but moderate MSV optimizes behavioral engagement. This research advances the theoretical understanding of short video engagement and introduces a robust computational tool for short video research.
}
\keywords{Video-As-Data, Message Sensation Value, Short Video Analytics, Social Media Engagement}

\usepackage{tabularx}  
\usepackage{booktabs}  
\usepackage[mode=text]{siunitx} 
\usepackage{graphicx} 
\usepackage{csquotes}\MakeOuterQuote{"}  
\usepackage[hidelinks]{hyperref} 

\begin{document}
\maketitle



\section{Introduction}

Short video platforms have become important information sources, but how individuals process multimodal features remains unclear. Recent studies have been examining individual multimodal features in short videos, such as brightness \parencite{lu_unpacking_2023} and audio tempo \parencite{li_correction_2024}. However, since individuals process multimodal information collectively, not in isolation, existing findings on single multimodal features’ effects fall short of capturing the complex, integrated nature of multimodal information processing. This underscores a key research gap: A unified, theory-driven approach to understanding the holistic effect of multimodal features \parencite{lahat_multimodal_2015}.

Message Sensation Value (MSV) provides a useful theoretical framework for addressing this challenge. MSV quantifies the extent to which objectively measured message features and audiovisual content attract attention and elicit affective responses \parencite{palmgreen_sensation_1991}. MSV influences individual attention allocation by activating the neural system and triggering the sensory processing of information \parencite{mesulam_sensation_1998}. However, competing theoretical arguments and empirical evidence exist on how exactly MSV influences attention. Theories grounded in communication and cognitive science, such as the Activation Model of Information Exposure \parencite[AMIE;][]{donohew_activation_1980}, the Limited Capacity Model of Mediated Message Processing \parencite[LC4MP;][]{lang_limited_2000}, and the Elaboration Likelihood Model \parencite[ELM;][]{petty_elaboration_1986}, have argued for opposite mechanisms. AMIE and LC4MP propose that MSV functions as an attention-getter, whereas ELM argues for an attention-distracting role. Both arguments have received empirical support, and a more precise examination of the theoretical mechanism of MSV’s effects requires a reliable and scalable predictive model.

Grounded on the literature on MSV, Morgan and colleagues (2003) developed a measure of MSV by manually annotating formal (e.g., shot changes) and content (e.g., surprise ending) features. While foundational, this objective MSV measure that was contextualized in televised Public Service Announcements (PSAs) needs to be adapted to the contemporary media environment. The short video format is very different from PSA’s long video format, so many features selected based on PSAs may not be relevant (e.g., formats unexpected in PSAs) or be quantifiable by existing computational models (e.g., surprise end). Distinctive from PSAs, short video formats are used across a broader range of content beyond science and health topics, created and distributed by both ordinary users and institutions, consumed within algorithm-curated video feeds rather than broadcast television, and engaged with interactive social media features (e.g., likes, comments, shares) beyond viewing alone. The combination of content diversity \parencite{shutsko_user-generated_2020}, algorithm curation \parencite{zannettou_analyzing_2024}, and real-time engagement features \parencite[e.g., Danmu in][]{wang2022community} makes short video processing and engagement dynamics much more complex than the traditional PSA engagement processes from the mass media era. Thus, it necessitates the reinvestigation of MSV and its influences on sensational experiences and behavioral engagement. Further, on the methodological front, the past research on MSV used traditional content analyses involving labor-intensive and manual annotation processes. The approach of using binary annotation of features also oversimplifies continuous audiovisual dynamics. These limitations highlight the need for a robust and scalable model that automatically predicts the holistic sensory impact of multimodal features.

The present study addresses this gap by developing a computational model that (a) quantifies the MSV of multimodal features in short videos and (b) predicts viewer engagement. Short videos tend to be designed and prioritized by platforms to attract more rapid, sensation-driven attention. This focus aligns with the literature on MSV and sensory message processing since attention is one of the primary responses elicited by MSV \parencite{donohew_attention_1994}. Unlike traditional PSA messages communicated one-way on TV, the engagement with short videos is more dynamic and collectively shaped by individual users and recommendation algorithms. We conceptualize video engagement along two dimensions, following the two-stage model of attention \parencite{epstein_quantifying_2022}. First, sensory engagement captures viewers’ immediate, affective attention to sensational videos, which is measured with Perceived Message Sensation Value (PMSV). It reflects an individual’s subjective perception and evaluation of a message in terms of the extent of emotional arousal, novelty, and dramatic impact \parencite{palmgreen_perceived_2002}. PMSV here serves as a subjective proxy for how sensational a video feels to viewers. Second, behavioral engagement represents viewers’ actions toward a video (e.g., likes, comments, shares) in real media environments. Engagement metrics reflect a higher level of engagement when viewers are activated to like, comment on, and share a video \parencite{epstein_quantifying_2022}. These two dimensions capture a fuller process of how short videos attract attention and translate sensory stimulation into behavior activation.

This study used (a) computational video analysis to extract 20 multimodal features from \textit{N} = 1200 short videos on Instagram Reels across eight critical societal issues and (b) machine learning models to pinpoint the combination of relevant multimodal features that best predict sensory engagement (i.e., PMSV; $N_{\text{participant}}$ = 1007). We focus on sensory engagement rather than behavioral engagement for model development because PMSV reflects human perceptions of a video’s capacity to elicit sensory experience, which is the most theoretically aligned and empirically measurable approximation of MSV. Specifically, PMSV measures how sensational a video is perceived by viewers, whereas MSV refers to the message properties that give rise to that perception. This choice is also supported by solid empirical evidence on a strong relationship between MSV and PMSV \parencite{morgan_associations_2003}. Therefore, using PMSV as an established measure of sensory experiences to back-develop the computational MSV model is both theoretically and empirically justified.

The contribution of this research is threefold. First, the findings contribute to the ongoing debates on the attentional mechanism of MSV by presenting correlational evidence of an inverted U-shaped curve between MSV and public engagement. This finding suggests that MSV may attract attention up to a point, highlighting the potential importance of the “middle ground” of MSV and advocating for a more granular examination of MSV in future research. Second, we provide a scalable and theory-informed computational model of MSV, which extends the original MSV measure beyond PSAs to the short-video context. This approach advances the theoretical understanding of multimodal message processing and provides a methodological framework for future computational analyses and model development. Lastly, we provide practical implications and actionable guidelines for content production on short video platforms.

\section{Message Sensation Value and Human Perceptions of It}
MSV quantifies the extent to which objectively measured multimodal features attract attention and elicit affective responses \parencite{palmgreen_sensation_1991}. By definition, MSV is a message-level construct: it captures the sensation-inducing potential embedded in a video’s multimodal features. In contrast, how humans perceive these features in terms of emotional arousal, novelty, and dramatic impact is captured through PMSV \parencite{palmgreen_perceived_2002}. PMSV is an observer-level construct, reflecting audience response rather than message properties. Therefore, MSV and PMSV represent distinct but related facets of a single underlying phenomenon: MSV concerns a video’s objective message properties, whereas PMSV concerns subjective audience response to those objective properties.

Distinguishing between MSV and PMSV clarifies how objective message attributes translate into human perceptions. Prior research showed that MSV and PMSV were moderately correlated \parencite{morgan_associations_2003}. This finding suggests that not all multimodal features may have meaningful contributions to PMSV or do so in varying directions. For example, shot changes and background music were unrelated to PMSV \parencite{morgan_associations_2003}. While visual complexity negatively predicted video engagement, audio loudness showed a positive association \parencite{lu_unpacking_2023}. Given the complex interplay among multimodal features, this study uses a machine learning-based approach to model a video’s sensation-inducing potential without treating MSV as a simple linear sum of individual features. Specifically, this approach examines how message-level multimodal features collectively predict viewer-level sensory engagement, operationalized as PMSV. 

\begin{quote}
    \textit{\textbf{RQ1.} How accurately can multimodal features in short videos predict sensory engagement?}
\end{quote}

Further, individual predispositions shape susceptibility to sensational content, especially the sensation-seeking tendency \parencite{zuckerman_sensation_2014} and age. Prior research found that high sensation-seekers prefer high-MSV content \parencite{everett_influences_1995, lorch_program_1994, wang_engaging_2015}. However, older adults generally experience lower arousal \parencite{smith_influences_2005} and reduced need for sensation \parencite{zuckerman_psychophysiology_1990}. To assess the stability of this computational model, we assess this model’s performance across different sensation-seeking and age groups.

\begin{quote}
    \textit{\textbf{RQ2.} How does the computational model of MSV perform across (a) sensation-seeking preference and (b) age?}
\end{quote}

\section{Behavioral Engagement with MSV}

MSV influences audience attention by activating the neural system and affecting attention allocation \parencite{mesulam_sensation_1998}. Viewers’ attention further drives downstream sensory processing and behavioral engagement with the media content. LC4MP explains that arousing and novel information receives more cognitive resources for processing, leading to increased attention and affective responses \parencite{lang_its_2005}. In addition, the AMIE model emphasizes the role of one’s optimal sensation level in conditioning the effect of MSV. Individuals pursue sensory experiences “in moderate amounts” \parencite[p.296]{donohew_activation_1980}, favoring messages with moderate MSV while avoiding understimulating or overstimulating messages. Both arguments treat message features as integral to message content, thereby influencing attention and preference.

However, ELM treats message content and message features as separate message components \parencite{petty_elaboration_1986}. MSV is considered as message features that activate peripheral processing \parencite{kang_attentional_2006}, while the main argument follows the central route, requiring more cognitive effort  \parencite{petty_elaboration_1986}. ELM highlights the attention-distracting role of MSV: as high MSV attracts attention to message features, fewer cognitive resources are allocated for message content processing.

Empirical evidence supports these different theoretical perspectives on the effect of MSV. On the one hand, research shows a dynamic interaction between MSV and individual sensation-seeking preferences. Physiological evidence showed that high-MSV videos activate the approach tendency among high sensation-seekers, which explains high sensation-seekers’ preference for high-MSV messages, while triggering the avoidance tendency among low sensation-seekers \parencite{everett_influences_1995, wang_engaging_2015}. These findings support the AMIE model’s notion that the effectiveness of high MSV lies in the alignment with optimal sensation levels \parencite{zuckerman_psychophysiology_1990}.

On the other hand, MSV shows a universal effect in enhancing message processing and engagement. For both high and low sensation-seekers, high-MSV PSAs were perceived as more effective \parencite{noar_assessing_2010}, effectively enhanced message processing and persuasion \parencite{stephenson_sensation_2001}, and improved message recall \parencite{niederdeppe_stylistic_2007}. Anti-smoking videos on YouTube with higher levels of MSV have received more views \parencite{paek_content_2010}. However, contradicting evidence exists to support the attention-distracting effect of MSV \parencite{kang_attentional_2006}. Neuroimaging data showed that low-MSV PSAs generated more brain activities and better message recall \parencite{seelig_low_2014}, challenging the notion that MSV facilitates message processing. A recent study on authoritative TikTok videos showed that MSV negatively predicted public engagement \parencite{zhang_factors_2023}. These mixed findings on MSV, attention, and subsequent message processing and engagement suggest a complex interplay between MSV and behavioral engagement. To further clarify this relationship, we examine how the computationally derived MSV would predict behavioral engagement across multiple contexts and short video platforms. 

\begin{quote}
    \textit{\textbf{RQ3.} How does the computational model of MSV predict behavioral engagement?}
\end{quote}

\begin{quote}
    \textit{\textbf{RQ4.} How does the computational model of MSV perform with unseen data?}
\end{quote}

\section{Methods}

\begin{figure}[h]
\centering
\includegraphics[width=0.95\columnwidth]{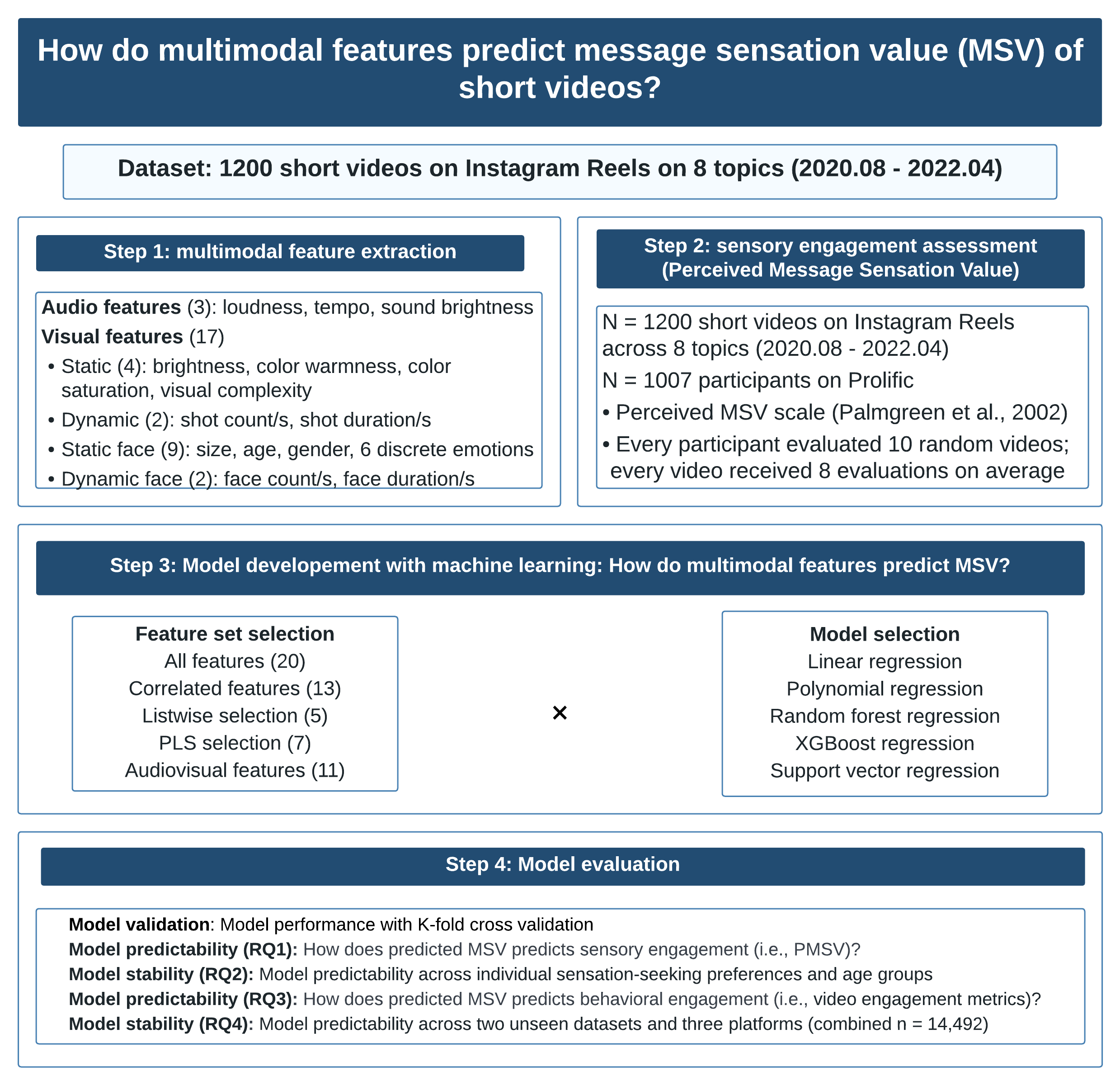}
\caption{Flow diagram of study procedures.}
\label{figure1}
\end{figure}

Relying on a dataset of 1200 short videos on Instagram Reels, we outline four key steps (Figure~\ref{figure1}) in developing a computational model of MSV: (a) extract multimodal features, (b) obtain human evaluations of PMSV in short videos, (c) develop a computational model using supervised machine learning, and (d) evaluate model stability and performance. We follow open science practice with open data, replication code, and Supplemental Information available at \href{https://osf.io/39nyj/}{\textbf{OSF}}.

\subsection{Data Collection and Sampling}

This dataset consists of 1200 short videos on Instagram Reels, covering eight critical societal issues (e.g., climate change, the Russian-Ukrainian conflict). Part of this dataset originated from \textcite{qian_convergence_2024}. They identified 20 hashtags and a pool of popular accounts (i.e., with more than 1 million followers) for each topic. We collected 39,634 videos published from August 2020 to April 2022 from these accounts with 4K Stogram, a data-extraction software \parencite{4k_stogram_4kstogram_2024}, and the metadata for 39,230 videos with Apify \parencite{apify_apify_2023}. The difference results from videos becoming unavailable or private during data collection. 

To increase the generalizability, we considered the influence of account popularity on MSV. We further identified eight ordinary accounts (i.e., with fewer than 50,000 followers) for each issue with the same hashtag. Within the same timeframe, we collected 8819 videos from these accounts and extracted metadata for 7699 videos.

Given the dataset scale and limited computational capacity, we used stratified sampling and created a sub-dataset of \textit{N} = 1200 videos for analysis. This approach considered both account popularity and issue type: among the 1200 videos sampled, 800 videos were published by popular accounts, and 400 were from ordinary accounts; 150 videos were sampled for each issue (100 videos from popular accounts and 50 videos from ordinary accounts). These videos were created by 153 unique Instagram accounts, with an average video length of 39.6 seconds (\textit{SD} = 57.0; \textit{Min} = 0.9; \textit{Max} = 559.7). See Table S1 in Supplemental Information for summary statistics.

\subsection{Multimodal Features in the Present Study}

The present study selects a set of multimodal features to (a) incorporate well-established audiovisual features in aesthetic literature, (b) capture both static aesthetics and dynamic motions in videos, and (c) align with the conceptual framework of MSV and current research on short video attributes and features. Further, we categorize visual features into static features captured on individual frames and dynamic features across frames. Table~\ref{tab:table1} details the definitions, feature extraction methods, and descriptive statistics of 20 multimodal features in the study.

\begin{table*}[t]
\centering
\caption{Description of multimodal features.}
\label{tab:table1}
\small

\begin{tabular}{p{.01\textwidth}
                p{.06\textwidth}
                p{.07\textwidth}
                p{.22\textwidth}
                p{.075\textwidth}
                p{.04\textwidth}
                p{.055\textwidth}
                p{.175\textwidth}}
\toprule
\# & Modality & Category & Multimodal feature & Range & Min & Max & M (SD) \\
\midrule

1  & \multirow{17}{*}{Visual}
   & \multirow{4}{*}{\shortstack{Static\\Visual}}
   & Brightness & 0, 255 & 4.49 & 250.80 & 121.03 (40.21) \\
2  &  &  & Color saturation & 0, 255 & 0.00 & 248.54 & 74.52 (36.51) \\
3  &  &  & Color warmness & 0, 1 & 0.00 & 0.89 & 0.24 (0.18) \\
4  &  &  & Visual complexity & 0, 8 & 0.15 & 7.76 & 6.47 (1.31) \\
\cmidrule(lr){3-8}

5  &  & \multirow{2}{*}{\shortstack{Dynamic\\Visual}}
   & Shot count/s & 0, $+\infty$ & 0.00 & 2.44 & 0.24 (0.29) \\
6  &  &  & Shot duration/s & 0, $+\infty$ & 0.01 & 3.32 & 0.50 (0.42) \\
\cmidrule(lr){3-8}

7  &  & \multirow{9}{*}{\shortstack{Static\\Face}}
   & Face age & 0, $+\infty$ & 4.39 & 74.33 & 34.29 (9.82) \\
8  &  &  & Face gender (female) & 0, 1 & 0.00 & 1.00 & 0.39 (0.33) \\
9  &  &  & Face size & 0, 1 & 0.00 & 0.89 & 0.10 (0.13) \\
10 &  &  & Disgust & 0, 100 & 0.00 & 75.50 & 5.77 (7.96) \\
11 &  &  & Anger & 0, 100 & 0.00 & 92.99 & 6.49 (10.30) \\
12 &  &  & Fear & 0, 100 & 0.00 & 97.20 & 4.41 (8.76) \\
13 &  &  & Happiness & 0, 100 & 0.00 & 100.00 & 23.83 (23.43) \\
14 &  &  & Sadness & 0, 100 & 0.00 & 96.06 & 9.36 (11.60) \\
15 &  &  & Surprise & 0, 100 & 0.00 & 99.99 & 11.28 (13.24) \\
\cmidrule(lr){3-8}

16 &  & \multirow{2}{*}{\shortstack{Dynamic\\Face}}
   & Face count/s & 0, $+\infty$ & 0.00 & 6.69 & 0.37 (0.59) \\
17 &  &  & Face duration/s & 0, 1 & 0.00 & 1.00 & 0.27 (0.33) \\
\midrule

18 & \multirow{3}{*}{Audio}
   & \multirow{3}{*}{Audio}
   & Loudness & 0, 1 & 0.00 & 0.47 & 0.05 (0.07) \\
19 &  &  & Tempo & 0, $+\infty$ & 0.00 & 215.33 & 118.42 (31.26) \\
20 &  &  & Sound brightness & 0, $+\infty$ & 0.00 & 4890.05 & 1919.06 (658.32) \\
\bottomrule
\end{tabular}
\end{table*}

\subsubsection{Visual Features}

\textbf{Static visual features.} This study investigates four basic visual features: (1) brightness, (2) saturation, (3) warmness, and (4) visual complexity, representing intrinsic color features \parencite{wyszecki_color_2000}. \textit{Brightness} is defined as the extent to which “a given visual stimulus appears to be more or less intense” \parencite[p.487]{wyszecki_color_2000}. \textit{Saturation} represents “the degree to which a chromatic stimulus differs from an achromatic stimulus regardless of their brightness” \parencite[p.487]{wyszecki_color_2000}. \textit{Color warmness} refers to the extent to which colors fall in the warm spectrum, such as red and orange \parencite{motoki_light_2019}. While discrete colors (e.g., red, green) capture categorical differences in color hue, color warmness can serve as a continuous, consistent representation of these perceptual variations \parencite{valdez_effects_1994}. Existing findings on specific color hues largely converge along the warm-cool continuum for explaining their results \parencite{bakhshi_red_2015, sharma_how_2024}, which provides a consistent way to quantify and interpret color perceptions. \textit{Visual complexity} is “the level of detail or intricacy” \parencite{shen_understanding_2022}. These color features influence affect, cognition, and behavior \parencite{elliot_color_2014}. For example, brightness elicits stronger physiological reactions \parencite{wilms_color_2018}, while warm colors increase social media engagement \parencite{yu_color_2021}. While high-level features such as visual composition can shape perceptions, this study focuses on first using low-level visual features to predict sensory experiences, which may have captured some variations introduced by higher-level compositional properties. For example, brightness and saturation influence aesthetic and naturalness perceptions indirectly through the mediation of higher-level factors like spatial arrangement or scene geometry \parencite{ibarra_image_2017}.

\textbf{Dynamic visual features.} This study included two dynamic features related to video motion–shot count per second and shot duration per second–as the number of shot changes and the shot duration that can captivate the audience’s attention \parencite{cutting_evolution_2016}. The original MSV measure quantifies the number of cuts in a given video, but this approach does not account for the confounder of video length variation. To address this confounding issue, we adjust for video length by calculating shot changes per second and the average duration of shots per second.

\textbf{Static and dynamic face features.} We included face features for their direct relevance to sensory activation: human faces can elicit immediate sensory and emotional responses \parencite{vermeulen_fast_2010}. Specifically, we include four static face features captured with sampled frames, including face size, face age, face gender, and facial expressions (i.e., disgust, anger, fear, happiness, sadness, and surprise). Lastly, we included two dynamic face features: the number of faces per second and the duration of faces per second. Although face features go beyond basic visual features, we included them because human faces can directly elicit sensory experiences through emotional contagion—participants mimic and experience similar emotions when watching short videos with various facial expressions \parencite{hess_facial_2001}. Similarly, the number and duration of faces are adjusted with the video length. 

\textbf{Static Visual Feature Extraction: frame sampling.} The OpenCV Library in Python \parencite{bradski_opencv_2000} was used to extract aesthetic visual features: brightness, color warmness, color saturation, and visual complexity. Given computational constraints, we used the equal interval frame sampling method with which a frame in every six frames is selected for analysis \parencite{wang_temporal_2016}. It uses average values of aesthetic features in the sampled frames to represent the overall aesthetics of a video. Despite equal interval frame sampling lacking flexibility in motion detection, it remains an effective technique for aesthetic feature estimation because aesthetic features are less dependent on keyframes and can be effectively captured with this approach.

\textbf{Static Visual Feature Extraction.} Frame-level brightness and color saturation are calculated by averaging the brightness and saturation values of every pixel in a frame, where brightness and saturation span from 0 to 255. Warmness is measured as the percentage of pixels with a hue between 0 (red) and 30 (yellow). Visual complexity is calculated as the diversity of pixel intensity in a frame \parencite{yang_using_2019}. It uses Shannon’s entropy formula to calculate the probability of occurrence of each unique pixel intensity level. It ranges from 0 to 8 (maximum), with a higher value representing a higher level of variability and complexity of visual patterns in a given frame.

\textbf{Dynamic Visual Feature Extraction: shot changes and face presence.} We used Google Cloud Video Intelligence API \parencite{google_cloud_video_2024} to extract the number of shots and the duration for each shot in a video. These two variables were adjusted by video length, thereby having the shot change counts and duration per second. This is because standardization by the time unit (a) enables accurate comparison across video lengths and (b) comes with a lower collinearity score than including video length as a separate variable (see Table S2 for details). Similarly, we extracted the number of faces and the average duration of faces, which were adjusted for video length.

\textbf{Static Face Feature Extraction.} We used Face++ \parencite{face_face_2023}, a face recognition software, to detect faces in an image and analyze various face attributes, including face size, estimated face age, estimated face gender, and six discrete facial expressions (i.e., disgust, anger, fear, happiness, sadness, and surprise). Face size is represented by the percentage of a face on a given frame, ranging from 0 to 1. The face age ranges from 0 to positive infinity. Face gender is calculated as the percentage of female faces in a given frame, ranging from 0 to 1. Lastly, facial expressions span from 0 to 100. Since Face++ analyzes images only, we adopted a frame-level approach by extracting face age and facial expressions of all identified faces in each frame and averaging these values to represent the overall face features in the video.

\subsubsection{Audio Features}

We include three basic auditory features: loudness, tempo, and sound brightness. Loudness is known as acoustic intensity, measuring sound energy \parencite{bannister_vigilance_2020}. Tempo describes the sound pace or speed, defined as “the rate at which underlying beat or pulse of music progresses” \parencite[p.568]{schubert_modeling_2004}. Sound brightness is quantified as spectral centroid to reflect sound frequency, with higher values signaling brighter sound \parencite{bannister_vigilance_2020}. These audio features are crucial in eliciting both individual emotional and physiological reactions and boosting engagement \parencite{trochidis_investigation_2011}. High-level audio features (e.g., background music and sound effects) in the objective MSV scale are not included because these fundamental audio features can capture the core properties of high-level features and represent the essence of sound.

\textbf{Audio Feature Extraction.} The Librosa Library in Python \parencite{mcfee_librosa_2015} was used to extract three audio features: loudness, tempo, and sound brightness. This tool has been widely used in audio-related studies across computer science and social science \parencite[e.g.,][]{lukito_audio-as-data_2024}. Loudness ranges from 0 to 1, indicating the energy of a sound wave. Tempo ranges from 0 to positive infinity, with a higher value indicating a faster pace. Sound brightness (measured as spectral centroid) spans from 0 to positive infinity, with a higher value representing a brighter sound and a lower value representing a darker sound.

\subsection{Short Video Engagement}

\textbf{Sensory Engagement with Short Videos.} To capture individual sensory engagement with short videos, we surveyed 1007 participants on Prolific in October 2023. All study procedures were approved by the university’s Institutional Review Board. Initially, 1086 U.S. adults were recruited. Responses from 79 participants were excluded due to failed attention checks, resulting in \textit{N} = 1,007 valid responses. Each participant evaluated the perceived Message Sensation Value for ten random short videos. The median time to complete the survey was 17.6 minutes. On average, each video received eight independent evaluations. Afterward, participants reported sensation-seeking preferences and demographics (see Table S3 for demographics). Specifically, PMSV was measured with an established PMSV scale \parencite{palmgreen_perceived_2002}. This instrument contains 17 items on a 7-point semantic differential scale, covering three dimensions: emotional arousal, dramatic impact, and novelty (\textit{n} = 10,069, \textit{Min} = 1, \textit{Max} = 7, \textit{M} = 3.65, \textit{SD} = 1.42, \textit{Cronbach’s $\alpha$} = .94). See Table S4 for the scale.

\textbf{Behavioral Engagement with Short Videos.} We operationalized the behavioral engagement with short videos as the sum of the number of views, likes, and comments because all engagement behaviors contribute to the visibility of short videos \parencite{kim_like_2017}. It is adjusted by a factor of 10,000 for simplicity.

\subsection{Measures}

\textbf{Sensation seeking.} The sensation-seeking preference was measured by a brief sensation-seeking scale \parencite{hoyle_reliability_2002}. Participants evaluate the extent to which they agree or disagree with eight statements on a 5-point scale (1: strongly disagree, 5: strongly agree) (\textit{N} = 1007, \textit{Min} = 1, \textit{Max} = 5, \textit{M} = 2.59, \textit{SD} = 0.85, \textit{Cronbach’s $\alpha$} = .81). See Table S5 for the scale.

\subsection{Model Development with Machine Learning}

We adopt a supervised machine learning approach that (a) systematically filters relevant multimodal features and (b) evaluates linear and non-linear machine learning models to identify the optimal model for predicting MSV with multimodal features.

\textbf{Feature Selection.} We use both data-driven and theory-driven approaches to identify five feature sets for further evaluation I am running a few minutes late; my previous meeting is running over. \parencite{maass_data-driven_2018}. Since not all multimodal features may directly contribute to MSV, excluding less relevant features enhances accuracy. Further, removing redundant features that are highly correlated can improve computational efficiency. The feature set 1 includes all 20 multimodal features. Table S6 lists the five distinct feature sets.

Three data-driven approaches are considered: correlation-based feature selection, stepwise feature selection, and partial linear squares regression (PLS) \parencite{jovic_review_2015}. First, feature set 2 includes 13 features with significant correlations with PMSV in Pearson’s correlation tests (see Table S7 for correlation). Second, by conducting forward, backward, and bidirectional stepwise linear regressions, feature set 3 emerged with five multimodal features that were consistently selected across three stepwise regressions. Third, we used PLS regression to identify multimodal features with significant effects on the underlying latent variable (i.e., PMSV) in a formative model. Multimodal features with absolute feature importance above 0.04 were considered impactful and included in this feature set (see Table S8 for feature importance). This approach results in feature set 4 with seven multimodal features.

Lastly, feature set 5 includes 11 audiovisual features, excluding facial attributes and facial expressions. This is because audiovisual features are universally present across video content, while facial attributes and expressions can be represented by low-level visual features. By zooming in on audiovisual features, this feature set presents high applicability and accuracy. Further, evaluating audiovisual features provides a baseline to understand low-level multimodal features’ contribution to MSV.

\textbf{Regression Model Selection.} We examine five regression models to account for both linear and non-linear relationships: linear regression, polynomial regression, random forest regression \parencite{segal_machine_2004}, XGBoost regression \parencite{chen_xgboost_2016}, as well as support vector regression \parencite{zhang_chapter_2020}.

\textbf{Evaluation metrics.} Model performance is assessed with two metrics: Mean Squared Error (MSE) and Akaike Information Criterion (AIC). MSE measures quantify the average squared difference between the estimated value and the actual value, with lower scores indicating higher accuracy and better model performance. AIC gauges model complexity and fit, with lower values representing higher levels of model parsimony and simplicity. We value both metrics to develop an MSV model with high accuracy and fit.

\textbf{Regression Model Data Preprocessing.} Before model development, we went through three steps for data preprocessing: (a) data imputation, (b) data standardization, and (c) train-test split. First, we used regression imputation to handle missing values \parencite{scheffer_dealing_2002}. Second, standardization was applied to all features on varying scales. Lastly, the data was randomly split into a training set (90\%, \textit{n} = 1080) for model development and a test set (10\%, \textit{n} = 120), which was not used in model training and reserved exclusively for testing. The variance of PMSV (\textit{Var} = 0.664) in the test set serves as the baseline MSE. It sets a straightforward threshold; any model yielding MSE above the baseline would be considered ineffective.

\textbf{Regression Model development.} Across five distinct feature sets and five regression models, we developed 25 models in total. Performance metrics (i.e., MSE and AIC) for all models are summarized in Table S9. All models achieve better performance than the baseline ($MSE_{\text{baseline}}$ = 0.664). Notably, the random forest regression model that uses 11 audiovisual features (\textit{MSE} = 0.448, \textit{AIC} = -74.31) demonstrated the top performance with the lowest MSE and the second lowest AIC score in the cohort.

In contrast to linear regression models with regression coefficients, random forest regression provides feature importance statistics that explain the contribution of each feature to model predictions (Table S10). The key multimodal features contributing to the model’s prediction are the duration of faces and shots per second, along with brightness, collectively accounting for 52\% importance. Using this model, we predicted the computational MSV (\textit{N} = 1200, \textit{Min} = 2.11, \textit{Max} = 4.86, \textit{M} = 3.51, \textit{SD} = 0.38) for further analysis.

\textbf{Model cross-validation.} In addition to the train-test split process in model development, we conducted a separate k-fold cross-validation across all 25 models \parencite{fushiki_estimation_2011}. Specifically, the dataset was divided into ten folds, with each fold serving once as a test set while models were trained on the other nine. This cross-validation procedure mitigates the potential biases that arise from the random train-test split in the training process and provides a robust evaluation of the model’s reliability in predicting MSV across different subsets of data. Table S11 summarizes the average MSE for every model in the 10-fold cross-validation. Notably, the random forest regression model that incorporates 11 audiovisual features consistently achieved the best performance with the lowest \textit{MSE} = 0.517. By comparing MSE across different models, we validated the stable performance of the best-performing model.

\subsection{Analysis Plan}

To answer RQ1 about how the computationally predicted MSV influences sensory engagement, we conducted a linear regression with the PMSV as the dependent variable and the computational MSV as the independent variable. This model incorporated covariates, including video posting time, video duration, video topic, and posting account popularity. As a robustness check, we conducted linear mixed effects models on each participant’s evaluation of the short videos ($n_{\text{evaluation}}$ = 10,070), with the computational MSV as the independent variable and the video evaluation by each participant as the dependent variable. The random effects of video presentation order and individual differences were controlled; six covariates (i.e., age, gender, race, education, income, and sensation seeking) were also included.

To answer RQ2 about the computational MSV model’s performance across (a) sensation-seeking and (b) age, we used similar linear regression models in RQ1 to assess the relationship between computational MSV and PMSV across different age groups and sensation-seeking preferences.

To answer RQ3 about how the computationally predicted MSV influences behavioral engagement, we conducted negative binomial regression models with video engagement as the dependent variable and the computational MSV as the independent variable. Since the literature suggests potential linear or non-linear relationships between MSV and engagement, we explored the computational MSV in both linear and polynomial terms.

Lastly, to answer RQ4 about the computational MSV model’s performance with unseen data, the same negative binomial regression models in RQ3 were conducted with video engagement as the dependent variable and the computational MSV as the independent variable for two new datasets.

\section{Results}

\subsection{Computational MSV Predicts Sensory Engagement}
\begin{figure}[h]
\centering
\includegraphics[width=0.95\columnwidth]{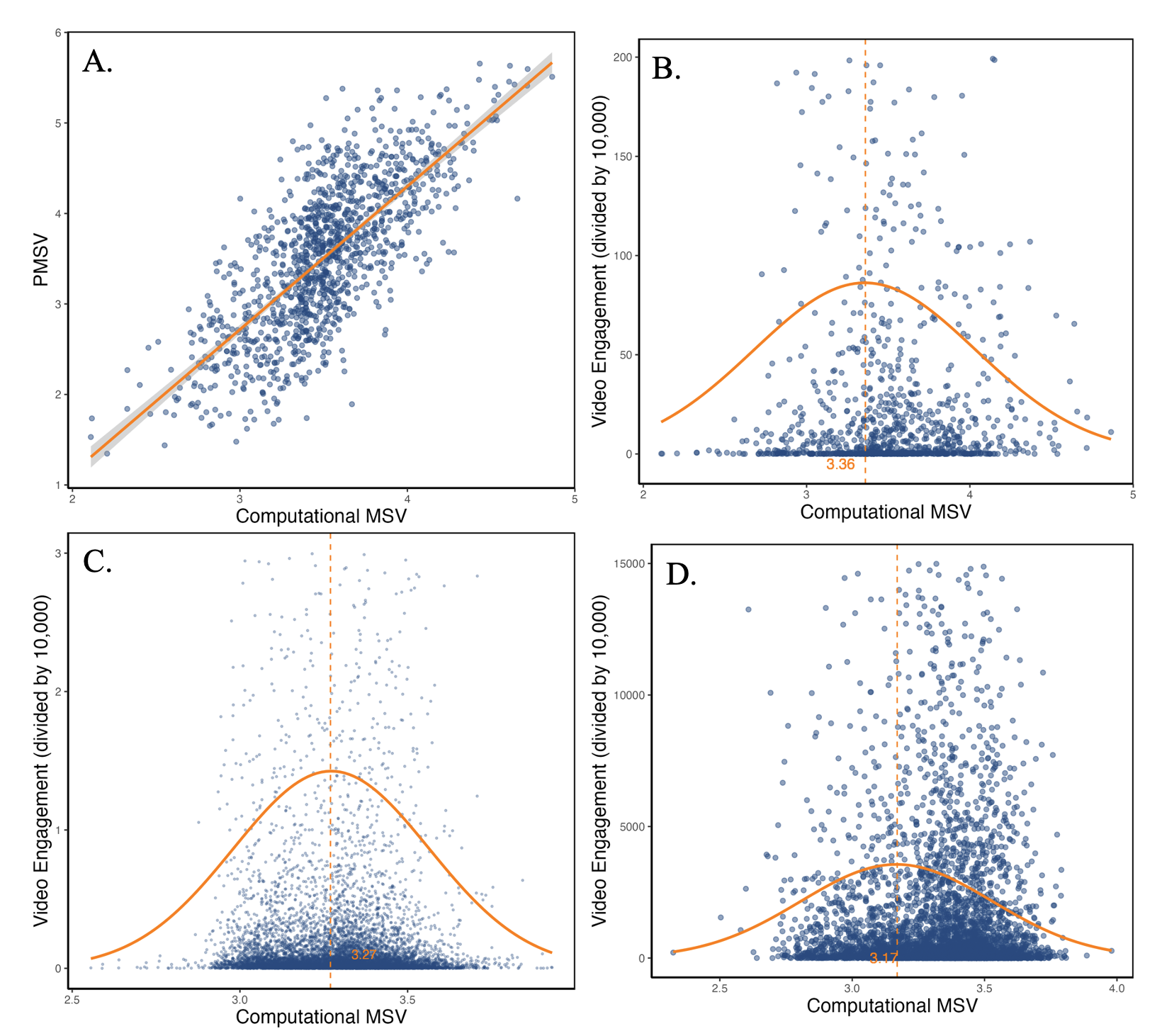}
\caption{The relationship between (A) the computational MSV and PMSV; (B) the computational MSV and video engagement, as illustrated in the orange curve. Video engagement is maximized with the optimal computational MSV = 3.36; (C) the computational MSV and video engagement for the dataset of science-related short videos (n = 10,000); and (D) the computational MSV and video engagement for the dataset of short videos for kids (n = 4,492).}
\label{figure2}
\end{figure}

To answer RQ1, we conducted two linear regression models on the association between the computational MSV and PMSV. See Figure~\ref{figure2} Panel A for the visualization and Table S12 for model details. The computational MSV was positively associated with PMSV (\textit{b} = 1.57, \textit{SE} = 0.04, \textit{95\% CI} = [1.49, 1.66], \textit{p} < .001), with or without adjustment for covariates. Further, linear mixed effects models supported this positive relationship at the level of individual video evaluation (\textit{b} = 1.57, \textit{SE} = 0.03, \textit{95\% CI} = [1.51, 1.63], \textit{p} < .001) (Table S13). These results collectively support that this computational MSV model can predict sensory engagement with short videos such that short videos with higher levels of MSV are perceived as more sensational.

To answer RQ2, we conducted similar regression models with PMSV rated by individuals across (a) sensation-seeking preference and (b) age. Specifically, we divided the data based on median sensation-seeking and age and examined the association between the computational MSV and PMSV across different groups. Younger participants aged from 19 to 38 (\textit{n} = 503, \textit{M} = 29.40, \textit{SD} = 5.05); older participants aged from 39 to 83 (\textit{n} = 503, \textit{M} = 52.68, \textit{SD} = 9.37). High sensation-seekers have an average sensation-seeking of 3.30 (\textit{n} = 500, \textit{Min} = 2.63, \textit{Max} = 5.00, \textit{M} = 3.30, \textit{SD} = 0.54), compared with the average of 1.90 for low sensation-seekers (\textit{n} = 509, \textit{Min} = 1.00, \textit{Max} = 2.50, \textit{M} = 1.90, \textit{SD} = 0.43). Older participants (\textit{Min} = 1.18, \textit{Max} = 6.03, \textit{M} = 3.63, \textit{SD} = 0.98) tended to evaluate a video as more sensational than younger participants (\textit{Min} = 1.00, \textit{Max} = 6.47, \textit{M} = 3.43, \textit{SD} = 0.93; \textit{t} = 5.31, \textit{p} < .001). However, there was no significant difference in PMSV between high sensation-seekers (\textit{Min} = 1.00, \textit{Max} = 5.76, \textit{M} = 3.56, \textit{SD} = 0.93) and low sensation-seekers (\textit{Min} = 1.00, \textit{Max} = 6.82, \textit{M} = 3.49, \textit{SD} = 0.96; \textit{t} = 1.87, \textit{p} = .062).

Similar linear regression models were conducted. See Figure S1 for visualization and Tables S14--S17 for model details. The computational MSV remained a significant positive predictor of PMSV for low sensation-seekers (\textit{b} = 1.59, \textit{SE} = 0.06, \textit{95\% CI} = [1.47, 1.71], \textit{p} < .001), high sensation-seekers (\textit{b} = 1.54, \textit{SE} = 0.06, \textit{95\% CI} = [1.42, 1.66], \textit{p} < .001), younger participants (\textit{b} = 1.51, \textit{SE} = 0.06, \textit{95\% CI} = [1.39, 1.62], \textit{p} < .001), and older participants (\textit{b} = 1.60, \textit{SE} = 0.06, \textit{95\% CI} = [1.47, 1.72], \textit{p} < .001). These results suggest that the computational MSV consistently predicts sensory engagement across varied population groups.

\subsection{Computational MSV Predicts Behavioral Engagement}

To answer RQ3 on behavioral engagement, we fit negative binomial regression models to the computational MSV and video engagement (see Table S18 for details). In the linear term, the computational MSV positively predicted video engagement in the unadjusted Model (\textit{b} = -0.49, \textit{SE} = 0.16, \textit{95\% CI} = [-0.86, -0.12], \textit{p} = .002), while this relationship did not hold significance with covariates adjusted (\textit{b} = -0.11, \textit{SE} = 0.11, \textit{95\% CI} = [-0.38, 0.15], \textit{p} = .318). On the contrary, incorporating the computational MSV as a polynomial term revealed a consistent and significant relationship with video engagement (polynomial term: \textit{b} = -1.08, \textit{SE} = 0.25, \textit{95\% CI} = [-1.55, -0.55], \textit{p} < .001; linear term: \textit{b} = 7.25, \textit{SE} = 1.77, \textit{95\% CI} = [3.48, 10.73], \textit{p} < .001). See Figure~\ref{figure2} Panel B for visualization and Table S19 for model details.

Given the consistency of the results, the polynomial regression on the computational MSV demonstrated greater robustness, illustrating an inverted U-shaped relationship between computational MSV and video engagement. This finding suggests that short videos with moderate MSV were more likely to optimize behavioral engagement, while videos with lower or higher MSV were less activating. Specifically, we identified that the optimal level of computational MSV was approximately 3.36 to maximize video engagement (Figure~\ref{figure2} Panel B). The summary statistics of multimodal features at the optimal MSV level (computational value between 3.24 and 3.74) and below or above this range are presented in Table S20. Results showed that short videos with the optimal MSV were neither oversimplistic nor excessively fast-paced–these videos tended to display a moderate pace with moderate levels of audiovisual features. For example, a 60-second video with approximately 12 shots would be ideal, contrasting the high-MSV version with 24 shots on average, which may potentially distract attention with the rapid pace. Short videos with a moderate level of MSV are likely to achieve higher levels of engagement.

\subsection{The Computational MSV Model’s Performance with Unseen Data}

Lastly, to assess the stability of the computational MSV model with different video contexts and short video platforms (RQ4), we applied the same negative binomial regression models, with polynomial terms of the computational MSV as the independent variable and video engagement as the dependent variable, to two short video datasets on (a) science-related short videos on TikTok and (b) short videos recommended to children on TikTok, Instagram Reels, and YouTube Shorts (combined \textit{n} = 14,492) \parencite{hilbert_bigtech_2025, xue_catching_2025}. See Table S21 for dataset descriptions and summary statistics.

Similar inverted U-shaped relationships between the computational MSV and behavioral engagement were confirmed for both datasets (see Figure~\ref{figure2} Panels C and D for visualization and Table S22 for model details). We identified the optimal level of computational MSV to be 3.27 to maximize video engagement for science-related short videos and 3.17 for short videos recommended to kids, supporting the finding that short videos with moderate MSV tend to maximize video engagement. Given the consistency of the inverted U-shaped relationship across different datasets, we conclude that the computational model of MSV demonstrates stability.

\section{Discussion}

With short videos becoming crucial information hubs and formats in the current media environment, it is urgent to understand how multimodal features influence multimodal information processing and engagement. Grounded in the literature on Message Sensation Value, this study developed a theory-informed computational model of MSV, which advances both empirical and theoretical research on short videos and public communication. Most notably, this model broadens MSV’s scope beyond health-related PSAs, revitalizing this concept for emerging multimodal media environments. By linking multimodal features to sensory and behavioral engagement, this model provides generalizable insights into the sensory and behavioral impacts of MSV across diverse short-video contexts. Relying on a computational analysis of 15,692 short videos across three platforms, we theorize the holistic effects of short videos and develop a machine learning-based model of MSV with computationally extracted multimodal features. This model’s performance in predicting both sensory and behavioral engagement with short videos was evaluated across multiple contexts, platforms, and demographic groups.

This study reveals an intriguing pattern: while MSV was positively associated with sensory engagement (i.e., PMSV), it followed an inverted U-shaped curve with behavioral engagement. It means that short videos with higher MSV tend to elicit more sensory processing and heightened emotional responses; however, they can reduce behavioral engagement (e.g., likes, comments, shares) once MSV surpasses viewers’ optimal sensory range. This seemingly contradictory finding can be explained by the fact that sensory and behavioral engagement represent two different, sequential stages of engagement: the initial, immediate sensory stimulation followed by behavioral activation \parencite{epstein_quantifying_2022}. Prior research found that people prefer sensational content at first, but they would act on less sensational but more credible content in later stages \parencite{epstein_quantifying_2022}. Therefore, the impacts of MSV may vary across stages of engagement rather than exerting a uniform effect. Based on large-scale data analysis, this study provides empirical evidence that MSV may facilitate sensory engagement linearly but promote behavioral engagement only up to a point. This finding helps reconcile the previous mixed results on MSV’s effects by suggesting that these studies may focus on different stages of engagement. Further, while existing studies often dichotomize MSV, overlooking the middle spectrum \parencite[e.g.,][]{kang_attentional_2006}, our findings highlight the importance of the “middle ground” of MSV and advocate for a more granular examination in future research.

The study’s findings offer several significant theoretical and practical implications. First, this study advances the theoretical understanding of how sensory stimulation operates in contemporary media environments characterized by short-form videos. Extending the original theorization of MSV, our findings highlight the complex dynamics of multimodal features working synergistically in shaping viewer responses. First, many multimodal features were found to significantly predict sensory responses; however, not all features contribute meaningfully and equally to MSV, and those that do may interact in complex ways in shaping sensory processing. Importantly, given the highly interactive social media environment, the finding that MSV predicts sensory and behavioral engagement in distinct ways underscores a necessity to theorize MSV’s impact using a multi-stage process.

Second, methodologically, we introduce a scalable, data-driven approach for quantifying MSV using multimodal features. In this study, we examined how observable multimodal features systematically predict shared perceptions across viewers. Distinguishing message-level MSV and observer-level PMSV allows us to separate message properties from audience responses, enabling models that generalize across viewers and contexts rather than being tied to a specific set of subjective ratings. As more computational tools, including AI solutions on multimodal feature extraction and analyses, are built with increasing accessibility, we foresee that more research can follow suit to test additional features across more media content and channel contexts. This approach not only provides an automated, robust tool for MSV prediction in short video research but also informs future computational studies in revitalizing classic theoretical concepts in today’s media environments with computational approaches.

Lastly, by pinpointing that mid-level MSV maximizes engagement, this study provides initial observational evidence that may inform content production on short video platforms. This research demonstrates that short videos with an optimal level of MSV tend to have a balanced pace and appropriate audiovisual features that are neither oversimplified nor excessively sensational. We move beyond identifying individual multimodal features in isolation and provide a holistic and quantifiable framework for understanding and potentially improving user engagement.

\subsection{Limitations and Future Directions}

This study is not without limitations. First, one significant limitation revolves around the scope of multimodal features examined. Although this study delves into 20 multimodal features, it does not extend to high-level, sophisticated features such as compositional layout, key objects, or platform-specific features (e.g., video filters, AI-generated voiceover). Further, we captured visual motion with shot changes only, but visual motion extends beyond shot changes. More sophisticated features, such as person or object movement, should be incorporated in future research. Future research could explore a wider range of higher-level multimodal features \parencite[e.g.,][]{peng_same_2018} to explore how semantic dimensions beyond sensory activation shape short video perceptions and engagement.

Second, it is worth noting that the study’s finding on the inverted U-shaped relationship between MSV and behavioral engagement remains correlational. More experimental research is needed to provide more causal evidence on MSV’s effects and examine the theoretical mechanism underlying such effects. Both correlational and causal evidence would contribute to a more robust conclusion on the impacts of MSV. In addition, we want to highlight that although this computational model of MSV was trained on PMSV, these two concepts should be treated as theoretically distinct. As an additional robustness check, we examined whether PMSV would predict engagement in patterns similar to computational MSV, and we found that it did not. Instead, PMSV showed a weaker and less consistent relationship with engagement. This pattern further supports the distinction between MSV as a message-level construct and PMSV as an observer-level response. It also points to a direction for future research: identifying additional observer-level indicators related to MSV that may shape downstream audience reactions.

Third, this study relies on a traditional machine-learning approach. With the increasing prevalence of multimodal large language models (LLMs) such as Veo 3 and Sora 2 \parencite{vallance_openai_2025}, future research could explore their potential in perceiving, interpreting, and analyzing multimodal features for more comprehensive multimodal information processing and understanding \parencite[e.g.,][]{shvetsova_howtocaption_2025}.

Fourth, this study did not specifically examine engagement within algorithm-curated short video feeds, even though behavioral engagement in terms of video engagement metrics reflects algorithmic influence to a certain extent \parencite{chan_ranking_2025}. On the one hand, how people engage with short videos can be substantially influenced by adjacent videos in the feed environment. On the other hand, how MSV in short videos may shape algorithms’ recommendation behaviors remains unknown. Future research should investigate the role of recommendation algorithms in moderating MSV’s impacts.

Fifth, this study employed proprietary tools for some feature extraction to ensure robustness and accuracy, but this approach may limit the reproducibility and generalizability of the proposed research methodology pipeline. Future studies are encouraged to validate and replicate the current approach with open-source alternatives as feature extraction tools continue to improve in performance and accessibility.

Lastly, this study is limited to the context of English short videos in the United States. The fact that MSV is a concept that varies across linguistic and cultural contexts \parencite{lu_sensation_2017} limits the generalizability of the study’s findings and the applicability of the computational model of MSV. To enhance the generalizability of the results, future research should replicate this study across a variety of cultural and linguistic settings and explore how the relationship between MSV and engagement changes with different media landscapes.



\printbibliography

@misc{4k_stogram_4kstogram_2024,
	title = {4kstogram: {Best} {IG} downloader by {URL}},
	shorttitle = {4kstogram},
	url = {https://www.4kstogram.com/},
	language = {en-US},
	urldate = {2024-03-27},
	journal = {4kstogram: Best IG downloader by URL},
	author = {4KStogram},
	year = {2024},
}

@misc{apify_apify_2023,
	title = {Apify: {Full}-stack web scraping platform},
	shorttitle = {Apify},
	url = {https://apify.com/},
	language = {en},
	urldate = {2023-12-06},
	journal = {Apify},
	author = {Apify},
	year = {2023},
}

@article{bakhshi_red_2015,
	title = {Red, {Purple} and {Pink}: {The} {Colors} of {Diffusion} on {Pinterest}},
	volume = {10},
	url = {https://doi.org/10.1371/journal.pone.0117148},
	doi = {10.1371/journal.pone.0117148},
	number = {2},
	journal = {PLOS ONE},
	publisher = {Public Library of Science},
	author = {Bakhshi, Saeideh and Gilbert, Eric},
	month = feb,
	year = {2015},
	pages = {e0117148},
}

@article{bannister_vigilance_2020,
	title = {A {Vigilance} {Explanation} of {Musical} {Chills}? {Effects} of {Loudness} and {Brightness} {Manipulations}},
	volume = {3},
	issn = {2059-2043},
	url = {https://doi.org/10.1177/2059204320915654},
	doi = {10.1177/2059204320915654},
	urldate = {2024-03-20},
	journal = {Music \& Science},
	publisher = {SAGE Publications Ltd},
	author = {Bannister, Scott},
	month = jan,
	year = {2020},
	pages = {2059204320915654},
}

@article{bradski_opencv_2000,
	title = {The {openCV} library.},
	volume = {25},
	number = {11},
	journal = {Dr. Dobb's Journal: Software Tools for the Professional Programmer},
	publisher = {Miller Freeman Inc.},
	author = {Bradski, Gary},
	year = {2000},
	pages = {120--123},
}

@misc{chan_ranking_2025,
	title = {The {Ranking} {Effect}: {How} {Algorithmic} {Rank} {Influences} {Attention} on {Social} {Media}},
	url = {https://arxiv.org/abs/2509.18440},
	author = {Chan, Jackie and Choi, Fred and Saha, Koustuv and Chandrasekharan, Eshwar},
	year = {2025},
	note = {\_eprint: 2509.18440},
}

@inproceedings{chen_xgboost_2016,
	address = {New York, NY, USA},
	series = {{KDD} '16},
	title = {{XGBoost}: {A} {Scalable} {Tree} {Boosting} {System}},
	isbn = {978-1-4503-4232-2},
	url = {https://doi.org/10.1145/2939672.2939785},
	doi = {10.1145/2939672.2939785},
	booktitle = {Proceedings of the 22nd {ACM} {SIGKDD} {International} {Conference} on {Knowledge} {Discovery} and {Data} {Mining}},
	publisher = {Association for Computing Machinery},
	author = {Chen, Tianqi and Guestrin, Carlos},
	year = {2016},
	pages = {785--794},
}

@article{cutting_evolution_2016,
	title = {The evolution of pace in popular movies},
	volume = {1},
	issn = {2365-7464},
	url = {https://doi.org/10.1186/s41235-016-0029-0},
	doi = {10.1186/s41235-016-0029-0},
	number = {1},
	journal = {Cognitive Research: Principles and Implications},
	author = {Cutting, James E.},
	month = dec,
	year = {2016},
	pages = {30},
}

@article{donohew_activation_1980,
	title = {An activation model of information exposure},
	volume = {47},
	issn = {0363-7751},
	url = {https://doi.org/10.1080/03637758009376038},
	doi = {10.1080/03637758009376038},
	number = {4},
	journal = {Communication Monographs},
	publisher = {Routledge},
	author = {Donohew, Lewis and Palmgreen, Philip and Duncan, Jack},
	month = nov,
	year = {1980},
	pages = {295--303},
}

@article{donohew_attention_1994,
	title = {Attention, {Need} for {Sensation}, and {Health} {Communication} {Campaigns}},
	volume = {38},
	issn = {0002-7642},
	url = {https://doi.org/10.1177/0002764294038002011},
	doi = {10.1177/0002764294038002011},
	number = {2},
	urldate = {2024-03-20},
	journal = {American Behavioral Scientist},
	publisher = {SAGE Publications Inc},
	author = {Donohew, Lewis and Palmgreen, Philip and Lorch, Elizabeth},
	month = nov,
	year = {1994},
	pages = {310--322},
}

@article{elliot_color_2014,
	title = {Color {Psychology}: {Effects} of {Perceiving} {Color} on {Psychological} {Functioning} in {Humans}},
	volume = {65},
	issn = {1545-2085},
	url = {https://www.annualreviews.org/content/journals/10.1146/annurev-psych-010213-115035},
	doi = {https://doi.org/10.1146/annurev-psych-010213-115035},
	number = {Volume 65, 2014},
	journal = {Annual Review of Psychology},
	publisher = {Annual Reviews},
	author = {Elliot, Andrew J. and Maier, Markus A.},
	year = {2014},
	note = {Type: Journal Article},
	pages = {95--120},
}

@misc{epstein_quantifying_2022,
	title = {Quantifying attention via dwell time and engagement in a social media browsing environment},
	author = {Epstein, Ziv and Lin, Hause and Pennycook, Gordon and Rand, David},
	year = {2022},
	note = {\_eprint: 2209.10464},
}

@article{everett_influences_1995,
	title = {Influences of {Sensation} {Seeking}, {Message} {Sensation} {Value}, and {Program} {Context} on {Effectiveness} of {Anticocaine} {Public} {Service} {Announcements}},
	volume = {7},
	issn = {1041-0236},
	url = {https://doi.org/10.1207/s15327027hc0703_3},
	doi = {10.1207/s15327027hc0703_3},
	number = {3},
	journal = {Health Communication},
	publisher = {Routledge},
	author = {Everett, Maureen W. and Palmgreen, Philip},
	month = jul,
	year = {1995},
	pages = {225--248},
}

@misc{face_face_2023,
	title = {Face++ {Cognitive} {Services}},
	url = {https://www.faceplusplus.com/},
	urldate = {2023-12-06},
	author = {Face++},
	year = {2023},
}

@article{fushiki_estimation_2011,
	title = {Estimation of prediction error by using {K}-fold cross-validation},
	volume = {21},
	issn = {1573-1375},
	url = {https://doi.org/10.1007/s11222-009-9153-8},
	doi = {10.1007/s11222-009-9153-8},
	number = {2},
	journal = {Statistics and Computing},
	author = {Fushiki, Tadayoshi},
	month = apr,
	year = {2011},
	pages = {137--146},
}

@misc{google_cloud_video_2024,
	title = {Video {AI}},
	url = {https://cloud.google.com/video-intelligence},
	language = {en},
	urldate = {2024-02-16},
	journal = {Google Cloud},
	author = {GoogleCloud},
	year = {2024},
}

@article{hess_facial_2001,
	title = {Facial mimicry and emotional contagion to dynamic emotional facial expressions and their influence on decoding accuracy},
	volume = {40},
	issn = {0167-8760},
	url = {https://www.sciencedirect.com/science/article/pii/S0167876000001616},
	doi = {10.1016/S0167-8760(00)00161-6},
	number = {2},
	journal = {International Journal of Psychophysiology},
	author = {Hess, Ursula and Blairy, Sylvie},
	month = mar,
	year = {2001},
	pages = {129--141},
}

@article{hilbert_bigtech_2025,
	title = {\#{BigTech} @{Minors}: social media algorithms have actionable knowledge about child users and at-risk teens},
	volume = {103},
	issn = {0736-5853},
	url = {https://www.sciencedirect.com/science/article/pii/S0736585325001030},
	doi = {10.1016/j.tele.2025.102341},
	journal = {Telematics and Informatics},
	author = {Hilbert, Martin and Cingel, Drew P. and Zhang, Jingwen and Vigil, Samantha L. and Shawcroft, Jane and Xue, Haoning and Thakur, Arti and Shafiq, Zubair},
	month = dec,
	year = {2025},
	pages = {102341},
}

@article{hoyle_reliability_2002,
	title = {Reliability and validity of a brief measure of sensation seeking},
	volume = {32},
	issn = {0191-8869},
	url = {https://www.sciencedirect.com/science/article/pii/S0191886901000320},
	doi = {10.1016/S0191-8869(01)00032-0},
	number = {3},
	journal = {Personality and Individual Differences},
	author = {Hoyle, Rick H and Stephenson, Michael T. and Palmgreen, Philip and Lorch, Elizabeth Pugzles and Donohew, R.Lewis},
	month = feb,
	year = {2002},
	pages = {401--414},
}

@article{ibarra_image_2017,
	title = {Image {Feature} {Types} and {Their} {Predictions} of {Aesthetic} {Preference} and {Naturalness}},
	volume = {Volume 8 - 2017},
	issn = {1664-1078},
	url = {https://www.frontiersin.org/journals/psychology/articles/10.3389/fpsyg.2017.00632},
	journal = {Frontiers in Psychology},
	author = {Ibarra, Frank F. and Kardan, Omid and Hunter, MaryCarol R. and Kotabe, Hiroki P. and Meyer, Francisco A. C. and Berman, Marc G.},
	year = {2017},
}

@inproceedings{jovic_review_2015,
	title = {A review of feature selection methods with applications},
	doi = {10.1109/MIPRO.2015.7160458},
	booktitle = {2015 38th {International} {Convention} on {Information} and {Communication} {Technology}, {Electronics} and {Microelectronics} ({MIPRO})},
	author = {Jović, A. and Brkić, K. and Bogunović, N.},
	year = {2015},
	pages = {1200--1205},
}

@article{kang_attentional_2006,
	title = {The {Attentional} {Mechanism} of {Message} {Sensation} {Value}: {Interaction} between {Message} {Sensation} {Value} and {Argument} {Quality} on {Message} {Effectiveness}},
	volume = {73},
	issn = {0363-7751},
	url = {https://doi.org/10.1080/03637750601024164},
	doi = {10.1080/03637750601024164},
	number = {4},
	journal = {Communication Monographs},
	publisher = {Routledge},
	author = {Kang, Yahui and Cappella, Joseph and Fishbein, Martin},
	month = dec,
	year = {2006},
	pages = {351--378},
}

@article{kim_like_2017,
	title = {Like, comment, and share on {Facebook}: {How} each behavior differs from the other},
	volume = {43},
	issn = {0363-8111},
	doi = {10.1016/J.PUBREV.2017.02.006},
	number = {2},
	journal = {Public Relations Review},
	publisher = {JAI},
	author = {Kim, Cheonsoo and Yang, Sung Un},
	month = jun,
	year = {2017},
	pages = {441--449},
}

@article{lahat_multimodal_2015,
	title = {Multimodal {Data} {Fusion}: {An} {Overview} of {Methods}, {Challenges}, and {Prospects}},
	volume = {103},
	doi = {10.1109/JPROC.2015.2460697},
	number = {9},
	journal = {Proceedings of the IEEE},
	author = {Lahat, Dana and Adali, Tülay and Jutten, Christian},
	year = {2015},
	pages = {1449--1477},
}

@article{lang_limited_2000,
	title = {The {Limited} {Capacity} {Model} of {Mediated} {Message} {Processing}},
	volume = {50},
	issn = {0021-9916},
	url = {https://doi.org/10.1111/j.1460-2466.2000.tb02833.x},
	doi = {10.1111/j.1460-2466.2000.tb02833.x},
	number = {1},
	urldate = {2023-07-25},
	journal = {Journal of Communication},
	author = {Lang, Annie},
	month = mar,
	year = {2000},
	pages = {46--70},
}

@article{lang_its_2005,
	title = {It's an {Arousing}, {Fast}-{Paced} {Kind} of {World}: {The} {Effects} of {Age} and {Sensation} {Seeking} on the {Information} {Processing} of {Substance}-{Abuse} {PSAs}},
	volume = {7},
	issn = {1521-3269},
	url = {https://doi.org/10.1207/S1532785XMEP0704_6},
	doi = {10.1207/S1532785XMEP0704_6},
	number = {4},
	journal = {Media Psychology},
	publisher = {Routledge},
	author = {Lang, Annie and Chung, Yongkuk and Lee, Seungwhan and Schwartz, Nancy and Shin, Mija},
	month = nov,
	year = {2005},
	pages = {421--454},
}

@article{li_correction_2024,
	title = {Correction by distraction: how high-tempo music enhances medical experts’ debunking {TikTok} videos},
	volume = {29},
	issn = {1083-6101},
	url = {https://doi.org/10.1093/jcmc/zmae007},
	doi = {10.1093/jcmc/zmae007},
	number = {5},
	urldate = {2024-09-03},
	journal = {Journal of Computer-Mediated Communication},
	author = {Li, Mengyu and Li, Gaofei and Yang, Sijia},
	month = sep,
	year = {2024},
	pages = {zmae007},
}

@article{lorch_program_1994,
	title = {Program {Context}, {Sensation} {Seeking}, and {Attention} to {Televised} {Anti}-{Drug} {Public} {Service} {Announcements}},
	volume = {20},
	issn = {0360-3989},
	url = {https://doi.org/10.1111/j.1468-2958.1994.tb00328.x},
	doi = {10.1111/j.1468-2958.1994.tb00328.x},
	number = {3},
	urldate = {2024-04-18},
	journal = {Human Communication Research},
	author = {Lorch, Elizabeth Pugzles and Palmgreen, Philip and Donohew, Lewis and Helm, David and Baer, Stagey A. and Dsilva, Margaret U.},
	month = mar,
	year = {1994},
	pages = {390--412},
}

@article{lu_sensation_2017,
	title = {Sensation {Seeking}, {Message} {Sensation} {Value}, and {Destinations}: {A} {Cross}-{Cultural} {Comparison}},
	volume = {41},
	issn = {1096-3480},
	url = {https://doi.org/10.1177/1096348014550872},
	doi = {10.1177/1096348014550872},
	number = {3},
	urldate = {2024-04-15},
	journal = {Journal of Hospitality \& Tourism Research},
	publisher = {SAGE Publications Inc},
	author = {Lu, Allan Cheng Chieh and Chi, Christina Geng-Qing and Lu, Carol Yi Rong},
	month = mar,
	year = {2017},
	pages = {357--383},
}

@article{lu_unpacking_2023,
	title = {Unpacking {Multimodal} {Fact}-{Checking}: {Features} and {Engagement} of {Fact}-{Checking} {Videos} on {Chinese} {TikTok} ({Douyin})},
	volume = {9},
	issn = {2056-3051},
	url = {https://doi.org/10.1177/20563051221150406},
	doi = {10.1177/20563051221150406},
	number = {1},
	urldate = {2023-08-10},
	journal = {Social Media + Society},
	publisher = {SAGE Publications Ltd},
	author = {Lu, Yingdan and Shen, Cuihua},
	month = jan,
	year = {2023},
	pages = {20563051221150406},
}

@article{lukito_audio-as-data_2024,
	title = {Audio-as-{Data} {Tools}: {Replicating} {Computational} {Data} {Processing}},
	volume = {12},
	issn = {2183-2439},
	doi = {https://doi.org/10.17645/mac.7851},
	journal = {Media and Communication},
	author = {Lukito, Josephine and Greenfield, Jason and Yang, Yunkang and Dahlke, Ross and Brown, Megan A. and Lewis, Rebecca and Chen, Bin},
	year = {2024},
}

@article{maass_data-driven_2018,
	title = {Data-{Driven} {Meets} {Theory}-{Driven} {Research} in the {Era} of {Big} {Data}: {Opportunities} and {Challenges} for {Information} {Systems} {Research}},
	volume = {19},
	issn = {1536-9323},
	shorttitle = {Data-{Driven} {Meets} {Theory}-{Driven} {Research} in the {Era} of {Big} {Data}},
	url = {https://aisel.aisnet.org/jais/vol19/iss12/1},
	doi = {10.17705/1jais.00526},
	number = {12},
	journal = {Journal of the Association for Information Systems},
	author = {Maass, Wolfgang and Parsons, Jeffrey and Purao, Sandeep and Storey, Veda and Woo, Carson},
	month = dec,
	year = {2018},
}

@inproceedings{mcfee_librosa_2015,
	title = {librosa: {Audio} and music signal analysis in python},
	volume = {8},
	booktitle = {Proceedings of the 14th python in science conference},
	author = {McFee, Brian and Raffel, Colin and Liang, Dawen and Ellis, Daniel P and McVicar, Matt and Battenberg, Eric and Nieto, Oriol},
	year = {2015},
	pages = {18--25},
}

@article{mesulam_sensation_1998,
	title = {From sensation to cognition.},
	volume = {121},
	issn = {0006-8950},
	url = {https://doi.org/10.1093/brain/121.6.1013},
	doi = {10.1093/brain/121.6.1013},
	number = {6},
	urldate = {2023-06-28},
	journal = {Brain},
	author = {Mesulam, M M},
	month = jun,
	year = {1998},
	pages = {1013--1052},
}

@article{morgan_associations_2003,
	title = {Associations {Between} {Message} {Features} and {Subjective} {Evaluations} of the {Sensation} {Value} of {Antidrug} {Public} {Service} {Announcements}},
	volume = {53},
	issn = {0021-9916},
	url = {https://doi.org/10.1111/j.1460-2466.2003.tb02605.x},
	doi = {10.1111/j.1460-2466.2003.tb02605.x},
	number = {3},
	urldate = {2023-07-29},
	journal = {Journal of Communication},
	publisher = {John Wiley \& Sons, Ltd},
	author = {Morgan, Susan E. and Palmgreen, Philip and Stephenson, Michael T. and Hoyle, Rick H. and Lorch, Elizabeth P.},
	month = sep,
	year = {2003},
	pages = {512--526},
}

@article{motoki_light_2019,
	title = {Light colors and comfortable warmth: {Crossmodal} correspondences between thermal sensations and color lightness influence consumer behavior},
	volume = {72},
	issn = {0950-3293},
	url = {https://www.sciencedirect.com/science/article/pii/S0950329318304385},
	doi = {10.1016/j.foodqual.2018.09.004},
	journal = {Food Quality and Preference},
	author = {Motoki, Kosuke and Saito, Toshiki and Nouchi, Rui and Kawashima, Ryuta and Sugiura, Motoaki},
	month = mar,
	year = {2019},
	pages = {45--55},
}

@article{niederdeppe_stylistic_2007,
	title = {Stylistic {Features}, {Need} for {Sensation}, and {Confirmed} {Recall} of {National} {Smoking} {Prevention} {Advertisements}},
	volume = {57},
	issn = {0021-9916},
	url = {https://doi.org/10.1111/j.1460-2466.2007.00343.x},
	doi = {10.1111/j.1460-2466.2007.00343.x},
	number = {2},
	urldate = {2024-03-22},
	journal = {Journal of Communication},
	author = {Niederdeppe, Jeff and Davis, Kevin C. and Farrelly, Matthew C. and Yarsevich, Jared},
	month = jun,
	year = {2007},
	pages = {272--292},
}

@article{noar_assessing_2010,
	title = {Assessing the {Relationship} {Between} {Perceived} {Message} {Sensation} {Value} and {Perceived} {Message} {Effectiveness}: {Analysis} of {PSAs} {From} an {Effective} {Campaign}},
	volume = {61},
	issn = {1051-0974},
	url = {https://doi.org/10.1080/10510970903396477},
	doi = {10.1080/10510970903396477},
	number = {1},
	journal = {Communication Studies},
	publisher = {Routledge},
	author = {Noar, Seth M. and Palmgreen, Philip and Zimmerman, Rick S. and Lustria, Mia Liza A. and Lu, Hung-Yi},
	month = jan,
	year = {2010},
	pages = {21--45},
}

@article{paek_content_2010,
	title = {Content analysis of antismoking videos on {YouTube}: message sensation value, message appeals, and their relationships with viewer responses},
	volume = {25},
	issn = {0268-1153},
	url = {https://doi.org/10.1093/her/cyq063},
	doi = {10.1093/her/cyq063},
	number = {6},
	urldate = {2024-03-22},
	journal = {Health Education Research},
	author = {Paek, Hye-Jin and Kim, Kyongseok and Hove, Thomas},
	month = dec,
	year = {2010},
	pages = {1085--1099},
}

@article{palmgreen_sensation_1991,
	title = {Sensation seeking, message sensation value, and drug use as mediators of {PSA} effectiveness.},
	volume = {3},
	issn = {1532-7027(Electronic),1041-0236(Print)},
	doi = {10.1207/s15327027hc0304_4},
	number = {4},
	journal = {Health Communication},
	publisher = {Lawrence Erlbaum},
	author = {Palmgreen, Philip and Donohew, Lewis and Lorch, Elizabeth P. and Rogus, Mary and Helm, David and Grant, Nancy},
	year = {1991},
	note = {Place: US},
	pages = {217--227},
}

@article{palmgreen_perceived_2002,
	title = {Perceived {Message} {Sensation} {Value} ({PMSV}) and the {Dimensions} and {Validation} of a {PMSV} {Scale}},
	volume = {14},
	issn = {1041-0236},
	url = {https://doi.org/10.1207/S15327027HC1404_1},
	doi = {10.1207/S15327027HC1404_1},
	number = {4},
	journal = {Health Communication},
	publisher = {Routledge},
	author = {Palmgreen, Philip and Stephenson, Michael T. and Everett, Maureen W. and Baseheart, John R. and Francies, Regina},
	month = oct,
	year = {2002},
	pages = {403--428},
}

@article{peng_same_2018,
	title = {Same {Candidates}, {Different} {Faces}: {Uncovering} {Media} {Bias} in {Visual} {Portrayals} of {Presidential} {Candidates} with {Computer} {Vision}},
	volume = {68},
	issn = {0021-9916},
	url = {https://doi.org/10.1093/joc/jqy041},
	doi = {10.1093/joc/jqy041},
	number = {5},
	urldate = {2025-10-17},
	journal = {Journal of Communication},
	author = {Peng, Yilang},
	month = oct,
	year = {2018},
	pages = {920--941},
}

@incollection{petty_elaboration_1986,
	address = {New York, NY},
	title = {The {Elaboration} {Likelihood} {Model} of {Persuasion}},
	isbn = {978-1-4612-4964-1},
	url = {https://doi.org/10.1007/978-1-4612-4964-1_1},
	doi = {10.1007/978-1-4612-4964-1_1},
	booktitle = {Communication and {Persuasion}: {Central} and {Peripheral} {Routes} to {Attitude} {Change}},
	publisher = {Springer New York},
	author = {Petty, Richard E. and Cacioppo, John T.},
	editor = {Petty, Richard E. and Cacioppo, John T.},
	year = {1986},
	pages = {1--24},
}

@article{qian_convergence_2024,
	title = {Convergence or divergence? {A} cross-platform analysis of climate change visual content categories, features, and social media engagement on {Twitter} and {Instagram}},
	volume = {50},
	issn = {0363-8111},
	url = {https://www.sciencedirect.com/science/article/pii/S036381112400033X},
	doi = {10.1016/j.pubrev.2024.102454},
	number = {2},
	journal = {Public Relations Review},
	author = {Qian, Sijia and Lu, Yingdan and Peng, Yilang and Shen, Cuihua (Cindy) and Xu, Huacen},
	month = jun,
	year = {2024},
	pages = {102454},
}

@article{scheffer_dealing_2002,
	title = {Dealing with missing data},
    journal = {Research Letters in the Information and Mathematical Sciences},
    volume = {3},
	publisher = {Massey University},
	author = {Scheffer, Judi},
	year = {2002},
    pages = {153-160}
}

@article{schubert_modeling_2004,
	title = {Modeling {Perceived} {Emotion} {With} {Continuous} {Musical} {Features}},
	volume = {21},
	issn = {0730-7829},
	url = {https://doi.org/10.1525/mp.2004.21.4.561},
	doi = {10.1525/mp.2004.21.4.561},
	number = {4},
	urldate = {2024-03-20},
	journal = {Music Perception},
	author = {Schubert, Emery},
	month = jun,
	year = {2004},
	pages = {561--585},
}

@article{seelig_low_2014,
	title = {Low {Message} {Sensation} {Health} {Promotion} {Videos} {Are} {Better} {Remembered} and {Activate} {Areas} of the {Brain} {Associated} with {Memory} {Encoding}},
	volume = {9},
	url = {https://doi.org/10.1371/journal.pone.0113256},
	doi = {10.1371/journal.pone.0113256},
	number = {11},
	journal = {PLOS ONE},
	publisher = {Public Library of Science},
	author = {Seelig, David and Wang, An-Li and Jaganathan, Kanchana and Loughead, James W. and Blady, Shira J. and Childress, Anna Rose and Romer, Daniel and Langleben, Daniel D.},
	month = nov,
	year = {2014},
	pages = {e113256},
}

@article{segal_machine_2004,
	title = {Machine {Learning} {Benchmarks} and {Random} {Forest} {Regression}},
	url = {https://escholarship.org/uc/item/35x3v9t4},
	language = {en},
	urldate = {2024-04-03},
	journal = {UCSF: Center for Bioinformatics and Molecular Biostatistics.},
	author = {Segal, Mark R.},
	month = apr,
	year = {2004},
}

@article{sharma_how_2024,
	title = {How {Visual} {Aesthetics} and {Calorie} {Density} {Predict} {Food} {Image} {Popularity} on {Instagram}: {A} {Computer} {Vision} {Analysis}},
	volume = {39},
	issn = {1041-0236},
	url = {https://doi.org/10.1080/10410236.2023.2175635},
	doi = {10.1080/10410236.2023.2175635},
	number = {3},
	journal = {Health Communication},
	publisher = {Routledge},
	author = {Sharma, Muna and Peng, Yilang},
	month = feb,
	year = {2024},
	pages = {577--591},
}

@article{shen_understanding_2022,
	title = {Understanding the {Effects} of {Visual} {Cueing} on {Social} {Media} {Engagement} {With} {YouTube} {Educational} {Videos}},
	volume = {65},
	doi = {10.1109/TPC.2022.3156225},
	number = {2},
	journal = {IEEE Transactions on Professional Communication},
	author = {Shen, Zixing and Tan, Songxin and Pritchard, Michael J.},
	year = {2022},
	pages = {337--350},
}

@inproceedings{shutsko_user-generated_2020,
	address = {Cham},
	title = {User-{Generated} {Short} {Video} {Content} in {Social} {Media}. {A} {Case} {Study} of {TikTok}},
	isbn = {978-3-030-49576-3},
	booktitle = {Social {Computing} and {Social} {Media}. {Participation}, {User} {Experience}, {Consumer} {Experience},  and {Applications} of {Social} {Computing}},
	publisher = {Springer International Publishing},
	author = {Shutsko, Aliaksandra},
	editor = {Meiselwitz, Gabriele},
	year = {2020},
	pages = {108--125},
}

@inproceedings{shvetsova_howtocaption_2025,
	address = {Cham},
	title = {{HowToCaption}: {Prompting} {LLMs} to {Transform} {Video} {Annotations} at {Scale}},
	isbn = {978-3-031-72992-8},
	booktitle = {Computer {Vision} – {ECCV} 2024},
	publisher = {Springer Nature Switzerland},
	author = {Shvetsova, Nina and Kukleva, Anna and Hong, Xudong and Rupprecht, Christian and Schiele, Bernt and Kuehne, Hilde},
	editor = {Leonardis, Aleš and Ricci, Elisa and Roth, Stefan and Russakovsky, Olga and Sattler, Torsten and Varol, Gül},
	year = {2025},
	pages = {1--18},
}

@article{smith_influences_2005,
	title = {Influences of {Age} on {Emotional} {Reactivity} {During} {Picture} {Processing}},
	volume = {60},
	issn = {1079-5014},
	url = {https://doi.org/10.1093/geronb/60.1.P49},
	doi = {10.1093/geronb/60.1.P49},
	number = {1},
	urldate = {2024-04-16},
	journal = {The Journals of Gerontology: Series B},
	author = {Smith, Darin P. and Hillman, Charles H. and Duley, Aaron R.},
	month = jan,
	year = {2005},
	pages = {P49--P56},
}

@article{stephenson_sensation_2001,
	title = {Sensation seeking, perceived message sensation value, personal involvement, and processing of anti-marijuana {PSAs}},
	volume = {68},
	issn = {0363-7751},
	url = {https://doi.org/10.1080/03637750128051},
	doi = {10.1080/03637750128051},
	number = {1},
	journal = {Communication Monographs},
	publisher = {Routledge},
	author = {Stephenson, Michael T. and Palmgreen, Philip},
	month = mar,
	year = {2001},
	pages = {49--71},
}

@inproceedings{trochidis_investigation_2011,
	address = {Padoue, Italy},
	title = {Investigation of the relationships between audio features and induced emotions in {Contemporary} {Western} music},
	url = {https://u-bourgogne.hal.science/hal-01882061},
	booktitle = {8th {Sound} and {Music} {Computing} {Conference} 2011},
	author = {Trochidis, Konstantinos and Delbé, Charles and Bigand, Emmanuel},
	month = jul,
	year = {2011},
}

@article{valdez_effects_1994,
	title = {Effects of color on emotions.},
	volume = {123},
	issn = {1939-2222(Electronic),0096-3445(Print)},
	doi = {10.1037/0096-3445.123.4.394},
	number = {4},
	journal = {Journal of Experimental Psychology: General},
	publisher = {American Psychological Association},
	author = {Valdez, Patricia and Mehrabian, Albert},
	year = {1994},
	note = {Place: US},
	pages = {394--409},
}

@article{vallance_openai_2025,
	title = {{OpenAI} video app {Sora} hits 1 million downloads faster than {ChatGPT}},
	url = {https://www.bbc.com/news/articles/crkjgrvg6z4o},
	language = {en-GB},
	urldate = {2025-10-27},
	journal = {BBC},
	author = {Vallance, Chris},
	month = oct,
	year = {2025},
}

@article{vermeulen_fast_2010,
	title = {Fast emotional embodiment can modulate sensory exposure in perceivers},
	volume = {3},
	issn = {null},
	url = {https://doi.org/10.4161/cib.3.2.10922},
	doi = {10.4161/cib.3.2.10922},
	number = {2},
	journal = {Communicative \& Integrative Biology},
	publisher = {Taylor \& Francis},
	author = {Vermeulen, Nicolas and Mermillod, Martial},
	month = mar,
	year = {2010},
	pages = {184--187},
}

@inproceedings{wang_temporal_2016,
	address = {Cham},
	title = {Temporal {Segment} {Networks}: {Towards} {Good} {Practices} for {Deep} {Action} {Recognition}},
	isbn = {978-3-319-46484-8},
	booktitle = {Computer {Vision} – {ECCV} 2016},
	publisher = {Springer International Publishing},
	author = {Wang, Limin and Xiong, Yuanjun and Wang, Zhe and Qiao, Yu and Lin, Dahua and Tang, Xiaoou and Van Gool, Luc},
	editor = {Leibe, Bastian and Matas, Jiri and Sebe, Nicu and Welling, Max},
	year = {2016},
	pages = {20--36},
}

@article{wang2022community,
  title={Community-building on Bilibili: the social impact of Danmu comments},
  author={Wang, Rui},
  journal={Media and Communication},
  volume={10},
  number={2},
  pages={54--65},
  year={2022}
}

@article{wang_engaging_2015,
	title = {Engaging {High}-{Sensation} {Seekers}: {The} {Dynamic} {Interplay} of {Sensation} {Seeking}, {Message} {Visual}-{Auditory} {Complexity} and {Arousing} {Content}},
	volume = {65},
	issn = {0021-9916},
	url = {https://doi.org/10.1111/jcom.12136},
	doi = {10.1111/jcom.12136},
	number = {1},
	urldate = {2023-07-30},
	journal = {Journal of Communication},
	author = {Wang, Zheng and Vang, Mao and Lookadoo, Kathryn and Tchernev, John M. and Cooper, Cody},
	month = feb,
	year = {2015},
	pages = {101--124},
}

@article{wilms_color_2018,
	title = {Color and emotion: effects of hue, saturation, and brightness},
	volume = {82},
	issn = {1430-2772},
	url = {https://doi.org/10.1007/s00426-017-0880-8},
	doi = {10.1007/s00426-017-0880-8},
	number = {5},
	journal = {Psychological Research},
	author = {Wilms, Lisa and Oberfeld, Daniel},
	month = sep,
	year = {2018},
	pages = {896--914},
}

@book{wyszecki_color_2000,
	title = {Color science: concepts and methods, quantitative data and formulae},
	volume = {40},
	publisher = {John wiley \& sons},
	author = {Wyszecki, Günther and Stiles, Walter Stanley},
	year = {2000},
}

@misc{xue_catching_2025,
	title = {Catching {Dark} {Signals} in {Algorithms}: {Unveiling} {Audiovisual} and {Thematic} {Markers} of {Unsafe} {Content} {Recommended} for {Children} and {Teenagers}},
	url = {https://arxiv.org/abs/2507.12571},
	author = {Xue, Haoning and Nishimine, Brian and Hilbert, Martin and Cingel, Drew and Vigil, Samantha and Shawcroft, Jane and Thakur, Arti and Shafiq, Zubair and Zhang, Jingwen},
	year = {2025},
	note = {\_eprint: 2507.12571},
}

@inproceedings{yang_using_2019,
	address = {New York, NY, USA},
	series = {{CHI} {EA} '19},
	title = {Using {Screenshots} to {Predict} {Task} {Switching} on {Smartphones}},
	isbn = {978-1-4503-5971-9},
	url = {https://doi.org/10.1145/3290607.3313089},
	doi = {10.1145/3290607.3313089},
	booktitle = {Extended {Abstracts} of the 2019 {CHI} {Conference} on {Human} {Factors} in {Computing} {Systems}},
	publisher = {Association for Computing Machinery},
	author = {Yang, Xiao and Ram, Nilam and Robinson, Thomas and Reeves, Byron},
	year = {2019},
	pages = {1--6},
}

@article{yu_color_2021,
	title = {Color and engagement in touristic {Instagram} pictures: {A} machine learning approach},
	volume = {89},
	issn = {0160-7383},
	url = {https://www.sciencedirect.com/science/article/pii/S0160738321000761},
	doi = {10.1016/j.annals.2021.103204},
	journal = {Annals of Tourism Research},
	author = {Yu, Joanne and Egger, Roman},
	month = jul,
	year = {2021},
	pages = {103204},
}

@inproceedings{zannettou_analyzing_2024,
	address = {New York, NY, USA},
	series = {{CHI} '24},
	title = {Analyzing {User} {Engagement} with {TikTok}'s {Short} {Format} {Video} {Recommendations} using {Data} {Donations}},
	isbn = {979-8-4007-0330-0},
	url = {https://doi.org/10.1145/3613904.3642433},
	doi = {10.1145/3613904.3642433},
	booktitle = {Proceedings of the 2024 {CHI} {Conference} on {Human} {Factors} in {Computing} {Systems}},
	publisher = {Association for Computing Machinery},
	author = {Zannettou, Savvas and Nemes-Nemeth, Olivia and Ayalon, Oshrat and Goetzen, Angelica and Gummadi, Krishna P. and Redmiles, Elissa M. and Roesner, Franziska},
	year = {2024},
}

@incollection{zhang_chapter_2020,
	title = {Chapter 7 - {Support} vector regression},
	isbn = {978-0-12-815739-8},
	url = {https://www.sciencedirect.com/science/article/pii/B9780128157398000079},
	doi = {10.1016/B978-0-12-815739-8.00007-9},
	booktitle = {Machine {Learning}},
	publisher = {Academic Press},
	author = {Zhang, Fan and O'Donnell, Lauren J.},
	editor = {Mechelli, Andrea and Vieira, Sandra},
	month = jan,
	year = {2020},
	pages = {123--140},
}

@article{zhang_factors_2023,
	title = {Factors influencing public engagement in government {TikTok} during the {COVID}-19 crisis},
	volume = {ahead-of-print},
	issn = {0264-0473},
	url = {https://doi.org/10.1108/EL-06-2023-0150},
	doi = {10.1108/EL-06-2023-0150},
	number = {ahead-of-print},
	urldate = {2024-03-20},
	journal = {The Electronic Library},
	publisher = {Emerald Publishing Limited},
	author = {Zhang, Wei and Yuan, Hui and Zhu, Chengyan and Chen, Qiang and Evans, Richard David and Min, Chen},
	month = jan,
	year = {2023},
}

@book{zuckerman_sensation_2014,
	title = {Sensation seeking (psychology revivals): {Beyond} the optimal level of arousal},
	publisher = {Psychology Press},
	author = {Zuckerman, Marvin},
	year = {2014},
}

@article{zuckerman_psychophysiology_1990,
	title = {The {Psychophysiology} of {Sensation} {Seeking}},
	volume = {58},
	issn = {0022-3506},
	url = {https://doi.org/10.1111/j.1467-6494.1990.tb00918.x},
	doi = {10.1111/j.1467-6494.1990.tb00918.x},
	number = {1},
	urldate = {2024-03-22},
	journal = {Journal of Personality},
	publisher = {John Wiley \& Sons, Ltd},
	author = {Zuckerman, Marvin},
	month = mar,
	year = {1990},
	pages = {313--345},
}

\end{document}